\pdfoutput=1
\documentclass[10pt,a4paper]{article}

\usepackage[a4paper, margin=2.5cm]{geometry}
\usepackage[T1]{fontenc}
\usepackage[utf8]{inputenc}
\usepackage{newtxtext, newtxmath}
\usepackage{microtype}
\usepackage[parfill]{parskip}

\usepackage{amsmath}
\usepackage{graphicx}
\usepackage{booktabs}
\usepackage{longtable}
\usepackage{multirow}
\usepackage{xcolor}
\usepackage{caption}
\usepackage{subcaption}
\usepackage{enumitem}
\usepackage{placeins}
\usepackage{wrapfig}
\usepackage{needspace}
\usepackage{tikz}
\usetikzlibrary{arrows.meta, positioning, fit, decorations.pathreplacing, calc}
\usepackage[authoryear, round, sort&compress]{natbib}
\usepackage[colorlinks=true, linkcolor=black, citecolor=black, urlcolor=black]{hyperref}

\definecolor{seldonaccent}{HTML}{4E29FF}
\setlist{itemsep=1pt, topsep=2pt}

\title{\vspace{-1.0cm}\textbf{Seldon TFM family: Seldon-Mini, Seldon and Seldon+}}
\author{Neuralk-AI Research Team}
\date{November 2026\\[0.6cm]\includegraphics[height=0.55cm]{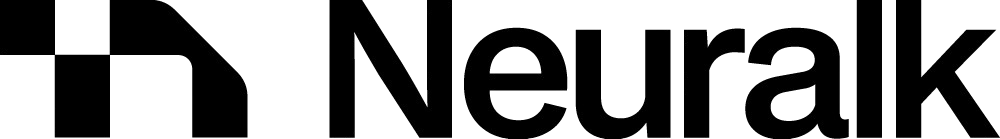}}

\begin{document}
\maketitle

\begin{abstract}
\noindent We introduce Seldon, a family of tabular foundation models.
Our classifier Seldon+ ranks first among 15 models on TabBench, our benchmark of 124 classification datasets, and on its variant with missing values.
In our own runs of public benchmarks, it also ranks first on BeyondArena IID classification and TabBench-Bio, and second on TabArena classification.
The open-weight models Seldon-Mini (27M parameters) and Seldon (213M parameters) both lie on the TabArena Pareto front of Elo against inference time: no entry is both more accurate and faster.
On TabArena, Seldon ranks fourth and Seldon-Mini seventh among 109 leaderboard entries, and on the IID track of BeyondArena, Seldon ranks second among 42 entries.
Seldon-Mini and Seldon handle both classification and regression, use an architecture similar to those of Kumo-Tabular, TabFM and TabPFN-3.5, and are trained on synthetic tables only.
\end{abstract}

\begin{table}[!h]
  \centering
  \footnotesize
  \begin{tabular}{l r l l}
    \toprule
    Model & Parameters (cls/reg) & Access & License \\
    \midrule
    \textbf{Seldon-Mini} & 27M / 28M & \href{https://huggingface.co/Neuralk-AI/seldon-mini}{\texttt{huggingface.co/Neuralk-AI/seldon-mini}} & Seldon Model License Agreement \\
    \textbf{Seldon}      & 213M / 215M & \href{https://huggingface.co/Neuralk-AI/seldon}{\texttt{huggingface.co/Neuralk-AI/seldon}} & Seldon Research and Evaluation License \\
    \textbf{Seldon+}     & -- & \href{https://prediction.neuralk.ai/register}{\texttt{prediction.neuralk.ai/register}} & -- \\
    \bottomrule
  \end{tabular}
\end{table}

\begin{figure}[!h]
  \centering
  \includegraphics[width=\textwidth]{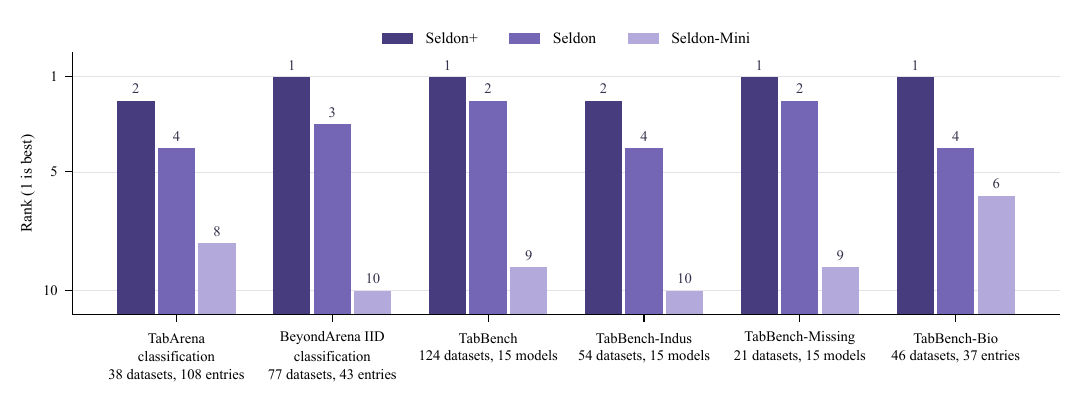}
  \caption{Overall rank of each Seldon model on each benchmark; a taller bar is a better rank. TabArena and BeyondArena IID: classification views, since Seldon+ is a classifier; official leaderboards, ranks among leaderboard entries, with the rank of Seldon+ from our own evaluation. TabBench suites: our evaluation service. TabBench-Bio: published leaderboard, reference setting. On TabArena, BeyondArena IID and TabBench-Bio, including Seldon+ moves every model below it down one place compared with the ranks quoted in the text.}
  \label{fig:ranks}
\end{figure}

\section{Introduction}

\begin{figure}[t]
  \centering
  \includegraphics[width=\textwidth]{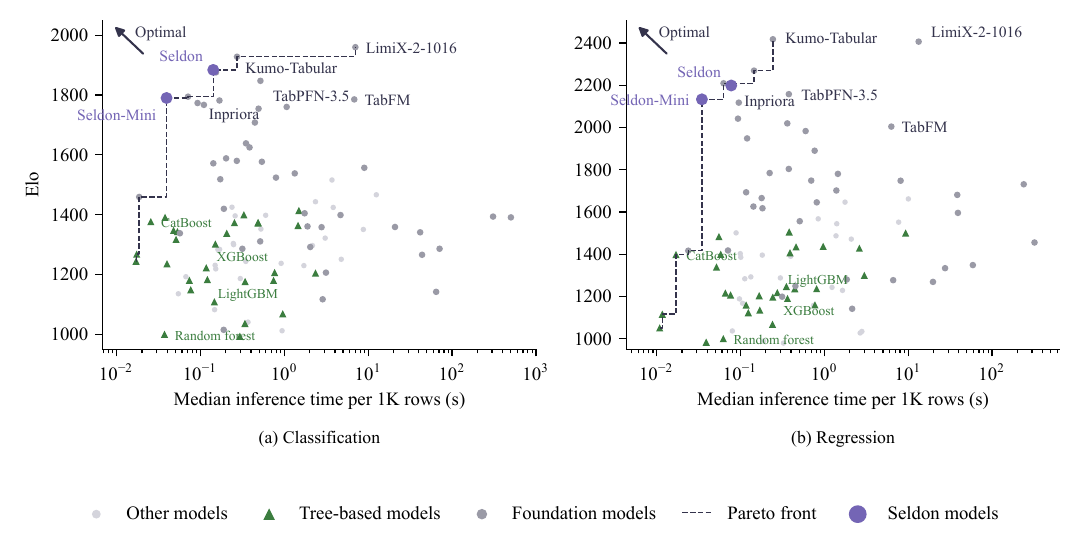}
  \caption{TabArena Elo against median inference time per 1{,}000 rows, for (a) classification (38 datasets) and (b) regression (13 datasets). Higher and further left is better. Each point is one entry of the official leaderboard (default view, 16 November 2026) with a reported time; times are measured on the hardware of each submission. Tree-based models are in green, and the default LightGBM, XGBoost, CatBoost and random forest are labelled. The y-axis starts at 950. The dashed line is the Pareto front. Seldon-Mini lies on the front in both panels, and Seldon in (a).}
  \label{fig:pareto}
\end{figure}

Gradient-boosted trees such as XGBoost \citep{chen2016xgboost} and CatBoost \citep{prokhorenkova2018catboost} have long been the default model for tabular data, but each new table requires its own training and hyperparameter tuning.
A tabular foundation model (TFM) needs neither.
It is pretrained once, typically on synthetic tables, and then makes predictions on a new table without being trained on it.
It reads the labelled rows of the table as context and predicts the labels of the remaining rows in a single forward pass, with no gradient step and no hyperparameter search.
Because much of the data behind decisions in finance, healthcare and industry is tabular, a model that is accurate on new tables without training or tuning could remove the per-task cycle of model selection and tuning, and bring strong predictive models to teams without machine-learning expertise.
TabPFN \citep{hollmann2023tabpfn, hollmann2025tabpfnv2} introduced this approach, and TabICL \citep{qu2025tabicl, qu2026tabiclv2} scaled it to larger tables.
TFMs now lead public leaderboards such as TabArena \citep{erickson2025tabarena}, ahead of tuned tree ensembles, with recent models including TabPFN-3.5 \citep{jaeger2026tabpfn35}, LimiX-2 \citep{zhang2026limix2}, TabFM \citep{kong2026tabfm} and Kumo-Tabular \citep{qu2026kumotabular}.
Yet these models differ widely in inference cost: among the ten best entries of TabArena, the median time to predict 1{,}000 rows ranges from under 0.1 to over 8 seconds (Figure~\ref{fig:pareto}).

This report presents Seldon, a family of three TFMs that succeeds an earlier Seldon model \citep{neuralk2026seldon}.
Two of them, Seldon-Mini (27M parameters) and Seldon (213M parameters), are released with open weights, each as a classifier and a regressor.
Like Kumo-Tabular, TabFM and TabPFN-3.5, they interleave column and row blocks, with a cell embedding close to that of TabFM, and they are trained from scratch on synthetic tables drawn from our own prior only.
The third, Seldon+, is a classifier served through the Neuralk API, for which we report results only.

We first evaluate the models on public leaderboards.
On TabArena, Seldon ranks fourth among 109 entries, and Seldon-Mini and Seldon both lie on the Pareto front of Elo against inference time, overall and on classification.
On the IID track of BeyondArena \citep{purucker2026beyondiid}, Seldon ranks second among 42 entries.
We also report the published leaderboard of the biomedical benchmark TabBench-Bio \citep{kreuer2026tabbenchbio}.
In our own runs, Seldon+ ranks first on BeyondArena IID classification and TabBench-Bio, and second on TabArena classification.

We then evaluate on our own benchmarks.
TabBench gathers 124 classification datasets, with two variants focused on industrial use cases (TabBench-Indus) and on missing values (TabBench-Missing).
On TabBench, Seldon+ and Seldon rank first and second among 15 models, and Seldon+ also ranks first on TabBench-Missing.
Breaking TFMs is a new suite of 10 synthetic probes, each isolating one failure mode of TFMs against a control table.

Figure~\ref{fig:ranks} summarises the rank of each Seldon model on every benchmark, and Figure~\ref{fig:pareto} places Seldon-Mini and Seldon against inference time on TabArena.
Section~\ref{sec:background} frames TFMs as prior-fitted networks and reviews their architectures, Section~\ref{sec:seldon} describes the Seldon models, and Section~\ref{sec:results} reports the results.

Our contributions are as follows.
\begin{itemize}
  \item Two open-weight TFMs, Seldon-Mini (27M parameters) and Seldon (213M parameters), trained only on synthetic tables, that compete with the strongest tabular models. On TabArena, Seldon ranks fourth among 109 entries, and third on classification. Seldon-Mini ranks seventh with eight times fewer parameters. Both lie on the Pareto front of Elo against inference time: no entry is both more accurate and faster.
  \item Results that hold beyond TabArena. Seldon ranks second among 42 entries on the IID track of BeyondArena. On TabBench, our suite of 124 classification datasets, the API model Seldon+ ranks first and Seldon second among 15 models. Seldon+ also ranks first on BeyondArena IID classification and TabBench-Bio in our own runs. On TabArena, BeyondArena and TabBench alike, Seldon wins more than half of its comparisons against at least four of six leading TFMs.
  \item Breaking TFMs, a suite of 10 synthetic probes that each isolate one failure mode of TFMs against a control table, with known Bayes-optimal accuracy. Seldon-Mini ranks first overall, and Seldon and Seldon-Mini are two of the only three models that solve the parity of eight bits.
\end{itemize}

\section{Background}
\label{sec:background}

Seldon-Mini and Seldon are trained only on synthetic tables, and their architecture builds on TabICL and on more recent TFMs that interleave column and row blocks.
Section~\ref{sec:bg-pfn} recalls why training on synthetic tables transfers to real ones, Section~\ref{sec:bg-prior} reviews the priors and training data of recent TFMs, and Section~\ref{sec:bg-arch} reviews their architectures.

\begin{figure}[t]
  \centering
  \resizebox{\textwidth}{!}{\input{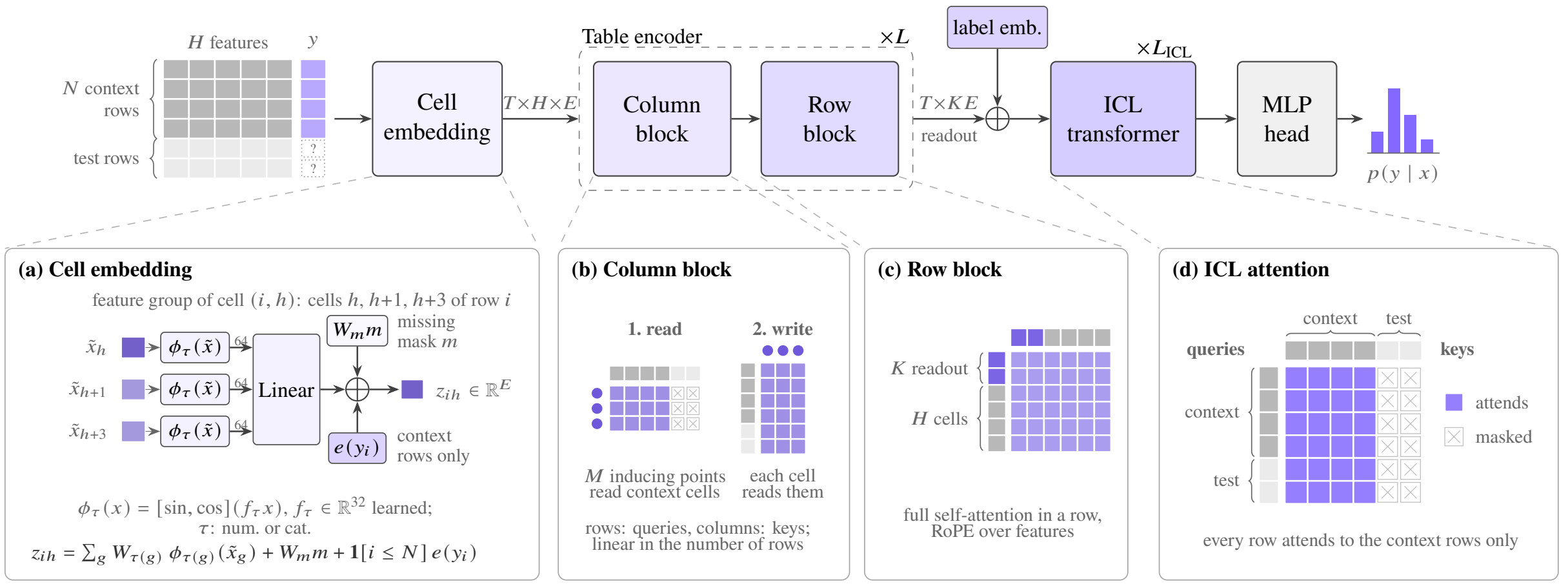}}
  \caption{Architecture of Seldon-Mini and Seldon. Top: a table of $T$ rows and $H$ features, whose $N$ context rows carry a label, goes through the cell embedding, $L$ layers of the table encoder, the readout, $L_{\mathrm{ICL}}$ ICL blocks and an MLP head, which returns class logits or 999 quantiles. Bottom, one panel per stage. (a) A cell is embedded from its feature group: itself and the cells at offsets 1 and 3 of its row. Each value, with a missing value replaced by the context mean of its feature, goes through learned Fourier features, with frequencies specific to numerical and categorical features. One linear map, with weights specific to numerical and categorical features, projects the three results to width $E$. A linear map of the group's missing mask is added, and so is the label embedding for a context row. In (b) to (d), each attention matrix has queries in rows and keys in columns. (b) In a column block, $M$ inducing points read the context cells of the column, then every cell of the column reads the inducing points. (c) In a row block, the $K$ readout tokens and the $H$ cells of the row attend to each other. (d) In an ICL block, every row attends to the context rows only. }
  \label{fig:arch}
\end{figure}

\subsection{Tabular Foundation Models as Prior-Fitted Networks}
\label{sec:bg-pfn}

TabPFN \citep{hollmann2023tabpfn} and most of its successors, Seldon included, are prior-fitted networks (PFNs) \citep{muller2022pfn}.
Let $D = \{(x_i, y_i)\}_{i=1}^{N}$ be the labelled context rows of a table and $x^\star$ a test row.
Let $\phi$ be a data-generating process drawn from a prior $p(\phi)$.
The Bayes-optimal predictive distribution is the posterior predictive
\begin{equation}
  p(y^\star \mid x^\star, D) = \int p(y^\star \mid x^\star, \phi)\, p(\phi \mid D)\, d\phi .
\end{equation}
This integral is intractable for any realistic prior.
A PFN $q_\theta$ approximates it instead.
It is trained on tables sampled from the prior, to minimise
\begin{equation}
  \mathcal{L}(\theta) = \mathbb{E}_{\phi \sim p(\phi),\; (D, x^\star, y^\star) \sim \phi} \big[ -\log q_\theta(y^\star \mid x^\star, D) \big].
  \label{eq:pfn-loss}
\end{equation}
At inference, the network conditions on the context rows of a real table and needs no gradient step.
It works well on a real table to the extent that this table resembles those drawn from the prior.

\subsection{Priors and Training Data}
\label{sec:bg-prior}

The prior determines which tables a PFN learns to handle. Recent TFMs differ in how they build it, and some add real tables to their training data.

\textbf{Structural causal models.} The prior of TabPFN \citep{hollmann2023tabpfn, hollmann2025tabpfnv2} samples a structural causal model (SCM): a random directed acyclic graph in which each node is a noisy random function of its parents.
The features and the target are read from a subset of the nodes, and the target is discretised into classes for classification.
Varying the graph, the functions and the noise yields tables with diverse feature interactions and dependence structures.

\textbf{Mixtures of priors.} TabICL \citep{qu2025tabicl} complements this prior with SCMs whose functions are tree ensembles, which produce the axis-aligned, piecewise-constant relations that gradient-boosted trees model well.
Mitra \citep{zhang2025mitra} studies mixtures of synthetic priors more systematically, and finds that the diversity and the distinctiveness of the priors in a mixture both matter for the performance of the resulting model on real tables.

\textbf{Real tables.} Other TFMs use real data.
TabDPT \citep{ma2025tabdpt} is pretrained on a collection of real tables with self-supervised objectives, and Real-TabPFN \citep{garg2025realtabpfn} continues the pretraining of TabPFN v2 on curated real-world tables.
Real data brings distributions that a synthetic prior may miss, but it limits the number of distinct training tables and must be kept disjoint from the evaluation benchmarks.

\textbf{Training on larger tables.} The cost of training grows with the number of rows of the tables, so several TFMs are trained on small tables first.
TabICL, for instance, trains in successive stages of increasing table size.

Seldon-Mini and Seldon follow the synthetic route: they are trained on an SCM prior only, without real tables (Section~\ref{sec:seldon}).

\subsection{Tabular Foundation Model Architectures}
\label{sec:bg-arch}

TFM architectures differ mainly in how they attend over the rows and the features of a table, and in how their cost grows with the size of the table.

\textbf{Cell-level attention.} TabPFN v2 \citep{hollmann2025tabpfnv2} represents each cell as a token and alternates attention across the features of a row and across the rows of a feature.
Every cell thus interacts with the other cells of its row and of its column, but the cost grows quadratically with the number of rows, for every feature.

\textbf{Column, row and ICL stages.} TabICL \citep{qu2025tabicl} reduces this cost by splitting the model into three stages.
A column stage embeds each cell with a set transformer over the column, with inducing points \citep{lee2019settransformer}.
A row stage then aggregates the cells of a row into a few readout tokens.
Finally, an in-context learning (ICL) stage attends from each test row to the context rows, over one vector per row instead of one per cell.

\textbf{Interleaved column and row blocks.} In TabICL, the column and row stages each run once.
Kumo-Tabular \citep{qu2026kumotabular}, TabFM \citep{kong2026tabfm} and TabPFN-3.5 \citep{jaeger2026tabpfn35} instead interleave them layer by layer: each layer applies a column block and a row block.
TabFM also embeds each cell with Fourier features of a group of features.

Seldon-Mini and Seldon use interleaved column and row blocks, with a cell embedding close to that of TabFM (Section~\ref{sec:seldon}).

\FloatBarrier
\section{Seldon}
\label{sec:seldon}

\subsection{Architecture}

Seldon-Mini has the size of Kumo-Tabular-Small, and Seldon the size of Kumo-Tabular, the large configuration \citep{qu2026kumotabular}.
Unlike the released Kumo-Tabular checkpoints, their attention and feed-forward linear layers have no bias.
Figure~\ref{fig:arch} shows the architecture and Table~\ref{tab:sizes} gives the sizes.
A table has $T$ rows and $H$ features, and its first $N$ rows are the labelled context rows.
The other rows are test rows.

\textbf{Cell embedding.} Each cell is embedded with the two cells of its row at offsets one and three, taken cyclically over the features.
The three cells form its feature group.
Their values pass through learned Fourier features, with 32 frequencies per value, and a linear projection to width $E$.
Numerical and categorical features use separate frequencies and projections.
A missing value is replaced by the mean of its feature over the context rows; for a categorical feature, this is the mean of the category codes.
A learned embedding of the missingness pattern of the feature group is added to the cell, so the model knows which values were imputed.
The label of each context row is embedded and added to every cell of that row.

\textbf{Table encoder.} The table encoder interleaves the column and row stages over $L$ layers.
In the column block, $M$ learned inducing points attend to the context cells of the column.
Every cell of the column then attends to the inducing points.
The cost of the column block is linear in the number of rows.
In the row block, $K=4$ learned readout tokens and the $H$ cells of the row attend to each other, with rotary position embeddings.
The readout tokens skip the column blocks.
In the last row block, only the readout tokens are queries.
Their outputs, concatenated, form a row vector of width $K E$.

\textbf{ICL stage.} A second label embedding is added to the vectors of the context rows.
In each ICL block, every row attends to the context rows only.
Test rows do not attend to each other.
The prediction for a test row therefore does not depend on the other test rows.
In Seldon, the test rows attend with two key-value heads, shared by the 16 query heads as in grouped-query attention, while the context rows use all 16 heads.
In Seldon-Mini, all rows use all 8 heads.

\textbf{Blocks and head.} All blocks are pre-norm transformer blocks with RMSNorm, query-key normalisation and a feed-forward layer of twice the width.
The classification head returns logits for up to 10 classes.
Above 10 classes, an error-correcting output code turns the task into $\max(\lceil C/9 \rceil, 4 \lceil \log_{10} C \rceil)$ 10-way tasks, for $C$ classes.
Each task keeps 9 classes apart and merges the others into the tenth, and every class is kept apart in at least one task.
The code is drawn to maximise the smallest Hamming distance between classes.
The model runs once per task, and the score of a class averages its log probabilities over the tasks that keep it apart.
The regression head returns 999 quantiles of the standardised target.
The point prediction is their mean.

\begin{table}[t]
  \centering
  \caption{Sizes of the Seldon models. $L$: table encoder layers; $E$: cell width; $M$: inducing points; $K$: readout tokens. Parameter counts, rounded, are for the classifier and the regressor, which differ only in their label embeddings and heads. The configuration of Seldon+ is not disclosed.}
  \label{tab:sizes}
  \small
  \begin{tabular}{l rrrrrrr}
    \toprule
    Model & $L$ & $E$ & $M$ & $K$ & ICL blocks $\times$ width & ICL heads & Parameters \\
    \midrule
    Seldon-Mini & 4 & 128 & 128 & 4 & $12 \times 512$  & 8  & 27M / 28M \\
    Seldon      & 6 & 256 & 256 & 4 & $24 \times 1024$ & 16 & 213M / 215M \\
    Seldon+     & \multicolumn{7}{c}{not disclosed} \\
    \bottomrule
  \end{tabular}
\end{table}

\subsection{Prior and Training}
\label{sec:prior-training}

Seldon-Mini and Seldon are trained on synthetic tables only.
The tables are drawn from a prior over structural causal models (SCMs), in the line of the priors of TabPFN \citep{hollmann2023tabpfn} and TabICLv2 \citep{qu2026tabiclv2}.
The classifier and the regressor are trained separately, on classification and regression tables drawn from this prior.

\textbf{Prior.} Each training table is generated by a random SCM.
A random directed acyclic graph is drawn over latent nodes, and the value of each node is computed from those of its parents,
\begin{equation}
  z_j = f_j\big(z_{\mathrm{pa}(j)}\big) \quad \text{for a node with parents,} \qquad z_j = f_j(\varepsilon_j) \quad \text{for a root node,}
  \label{eq:scm}
\end{equation}
where $\varepsilon_j$ is random and each $f_j$ is a random function drawn from a broad family of linear, smooth and piecewise-constant functions.
The features and the target are read from the nodes: numerical columns directly, categorical columns and classification targets by discretising node values.
The prior varies the numbers of rows, features and classes (up to 10), the share of categorical features, the size and density of the graph and the functions on its edges, so that the training tables cover a wide range of dependence structures.
Tables whose target cannot be predicted from the features are discarded.

\textbf{Missing values.} Half of the training tables receive missing cells, under missing-completely-at-random, missing-at-random and missing-not-at-random mechanisms, with cells often missing jointly over groups of columns.

\textbf{Objective.} A training table of $T$ rows is split into $N$ context rows $D$ and $T - N$ query rows $Q$, with every class present in the context.
The classifier minimises the cross-entropy on the query rows, the empirical form of the PFN loss of Equation~\eqref{eq:pfn-loss}:
\begin{equation}
  \mathcal{L}_{\mathrm{cls}}(\theta) = -\frac{1}{|Q|} \sum_{t \in Q} \log q_\theta(y_t \mid x_t, D),
  \label{eq:loss-cls}
\end{equation}
averaged over the query rows of all tables in a batch.
The regressor predicts the quantiles $\hat q_{\alpha_k}(x_t)$ of the standardised target $\tilde y_t = (y_t - \mu_D) / \sigma_D$, where $\mu_D$ and $\sigma_D$ are the mean and standard deviation of the target over the context rows, at the levels $\alpha_k = k / 1000$ for $k = 1, \dots, 999$.
It minimises the pinball loss averaged over these levels,
\begin{equation}
  \mathcal{L}_{\mathrm{reg}}(\theta) = \frac{1}{999\, |Q|} \sum_{t \in Q} \sum_{k=1}^{999} \rho_{\alpha_k}\big(\tilde y_t - \hat q_{\alpha_k}(x_t)\big),
  \qquad \rho_\alpha(u) = u\,\big(\alpha - \mathbf{1}[u < 0]\big),
  \label{eq:loss-reg}
\end{equation}
a discretisation of the continuous ranked probability score.

\subsection{Inference}

Predictions average an ensemble of forward passes: 8 for the Seldon classifier and 4 for the Seldon-Mini classifier and both regressors.
Each member applies one of four numerical transforms (identity, Yeo-Johnson, robust scaling or a rank-Gaussian transform), a sign flip of half of the numerical features and a feature permutation.
For classification, each member also shifts the class labels cyclically.
Half of the members shuffle the codes of the categorical features.
Classification averages the logits of the members.
Regression sorts the quantiles of each member and combines them with a trimmed mean.

\subsection{Usage}

Seldon-Mini and Seldon are available as \texttt{SeldonClassifier} and \texttt{SeldonRegressor} in the \texttt{neuralk-oss} Python package, which can be installed from PyPI (\href{https://pypi.org/project/neuralk-oss/}{\nolinkurl{pypi.org/project/neuralk-oss}}).
Both follow the scikit-learn interface: \texttt{fit} takes the labelled rows that serve as context, and \texttt{predict} returns predictions for new rows, with the ensemble described above.
The weights are hosted on Hugging Face, in gated repositories, under two licences.
Seldon-Mini uses the Seldon Model License Agreement, which allows internal business use, whereas Seldon uses the Seldon Research and Evaluation License, which is limited to non-commercial research and evaluation.
Seldon+ is not released as weights and is accessed through the Neuralk API.

\FloatBarrier
\section{Results}
\label{sec:results}

\subsection{Evaluation Protocol}
\label{sec:protocol}

\textbf{Benchmarks.} We evaluate the Seldon models on two groups of benchmarks, summarised in Table~\ref{tab:benchmarks}.
The public benchmarks, TabArena \citep{erickson2025tabarena}, BeyondArena \citep{purucker2026beyondiid} and TabBench-Bio \citep{kreuer2026tabbenchbio}, are maintained by independent groups, and we report their published leaderboards (Section~\ref{sec:public}).
The Neuralk private benchmarks, TabBench and Breaking TFMs, are built and run by us, and have no public leaderboard (Section~\ref{sec:private}).
TabBench-Bio is unrelated to TabBench despite the similar name.
Seldon-Mini and Seldon appear on the public leaderboards.
Seldon+ is served only through the Neuralk API and is absent from them; where we report it on a public benchmark, the result comes from our own run of that benchmark and is marked as such.

\begin{table}[!h]
  \centering
  \caption{Benchmarks used in this report, and the source of the results of each Seldon model.}
  \label{tab:benchmarks}
  \small
  \begin{tabular}{@{}l l l l@{}}
    \toprule
    Benchmark & Datasets & Seldon-Mini, Seldon & Seldon+ \\
    \midrule
    \multicolumn{4}{@{}l}{\emph{Public benchmarks}} \\
    TabArena \citep{erickson2025tabarena} & 51 (38 cls., 13 reg.) & official leaderboard & our run, classification \\
    BeyondArena, IID track \citep{purucker2026beyondiid} & 103 (77 cls., 26 reg.) & official leaderboard & our run, classification \\
    TabBench-Bio \citep{kreuer2026tabbenchbio} & 46 & published leaderboard & our run, rank only \\
    \midrule
    \multicolumn{4}{@{}l}{\emph{Neuralk private benchmarks}} \\
    TabBench, -Indus, -Missing (App.~\ref{app:tabbench}) & 124, 54, 21 (cls.) & our evaluation service & our evaluation service \\
    Breaking TFMs (App.~\ref{app:breaking-tfms}) & 31 synthetic (cls.) & our evaluation service & not evaluated \\
    \bottomrule
  \end{tabular}
\end{table}

\textbf{Setup.} The Seldon models are evaluated as released, without hyperparameter tuning on any benchmark.
The TabArena and BeyondArena results are those of the official leaderboards as of 16 November 2026.
On these leaderboards, an entry is one configuration of a model: default, tuned, or tuned and ensembled.
Seldon and Seldon-Mini each have a single entry, the released model, and both entries are marked as verified.
LimiX-2 is the LimiX-2-1016 checkpoint on TabArena and BeyondArena, and the first LimiX-2 checkpoint on TabBench.
The TabBench results come from our evaluation service, with 5-fold cross-validation.

\textbf{Metrics.} TabArena and BeyondArena score binary classification with ROC AUC, multiclass classification with log loss and regression with RMSE.
TabBench scores every dataset with ROC AUC, and TabBench-Bio scores classification with macro-F1 and regression with RMSE.
TabArena, BeyondArena, TabBench and TabBench-Bio rank models with a Bradley--Terry Elo fitted on per-split comparisons \citep{bradley1952rank, erickson2025tabarena}.
The 95\% confidence intervals come from a bootstrap over datasets, with 200 resamples on TabArena and BeyondArena and 100 on TabBench.
The anchor is the default random forest at 1000 on TabArena, and the default XGBoost \citep{chen2016xgboost} at 1000 on BeyondArena.
TabBench has no random forest on its board, so its Elo is centred instead: its 12 external models average 1000.
Its Elo scale is therefore not comparable with the other two.
On TabBench, the external models are first rated among themselves, and each Seldon model is then rated against this fixed field.
The Seldon models are therefore not compared with each other in the Elo fit, and we compare them directly with win rates.
The win rate of model A against model B is the fraction of comparisons that A wins, with ties counted as one half: per split on TabArena and BeyondArena, with each dataset weighted equally, and per dataset on TabBench.
A model that fails on a dataset receives the score of a reference model: the default random forest on TabArena and BeyondArena, CatBoost \citep{prokhorenkova2018catboost} on TabBench.

\subsection{Public Benchmarks}
\label{sec:public}

\subsubsection{TabArena}

TabArena has 51 datasets, 38 for classification and 13 for regression.

\begin{figure}[t]
  \centering
  \includegraphics[width=\textwidth]{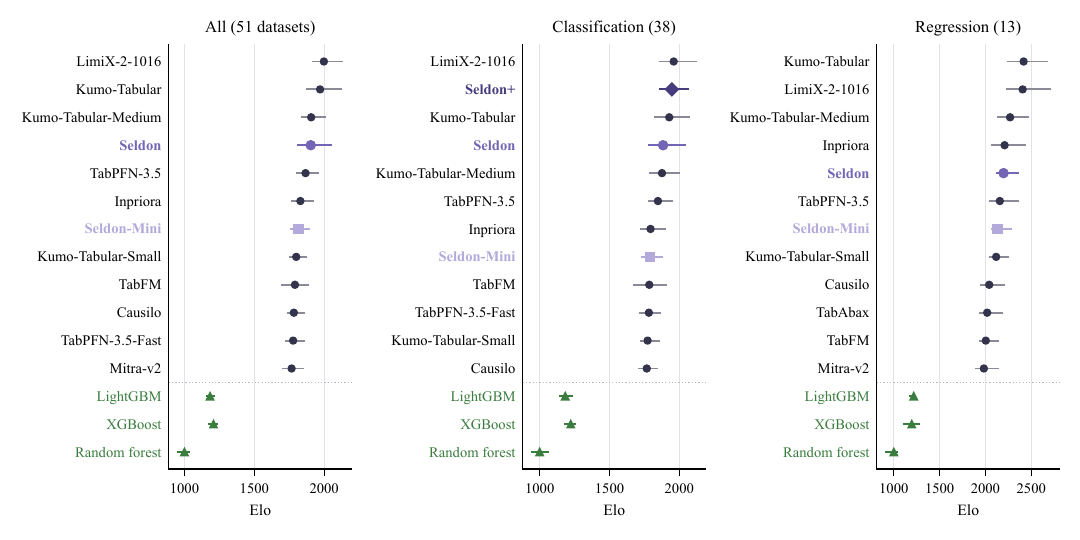}
  \caption{Official TabArena leaderboard (16 November 2026, default view: models only, imputed results included, all splits), top 12 entries of each task view. Elo with 95\% confidence interval; the default random forest is at 1000. Below the dotted line: the default LightGBM, XGBoost and random forest. All entries shown are default configurations. Seldon+ is not on the official leaderboard: it is added to the classification panel with its rating from our evaluation service, which scores every dataset with ROC AUC.}
  \label{fig:tabarena}
\end{figure}

\textbf{Seldon ranks fourth and Seldon-Mini seventh among 109 entries.}
Figure~\ref{fig:tabarena} gives the official leaderboard, with Seldon+ added to the classification panel.
Its 109 entries cover 59 models.
Seldon reaches an Elo of 1904 overall, 94 points below LimiX-2-1016 and 67 points below Kumo-Tabular.
Its 95\% interval overlaps those of the three entries above it.
On classification, Seldon ranks third, behind LimiX-2-1016 and Kumo-Tabular.
On regression, Seldon and Seldon-Mini rank fifth and seventh.
Seldon+ is a classifier, so it has no overall or regression rating.
On the 38 classification datasets, our evaluation service rates it at 1947$^{+126}_{-89}$, second among the 116 entries of its board, behind LimiX-2-1016 (1969) and ahead of Kumo-Tabular (1941).
The field of this board differs from that of the official classification view, which has 107 entries.
This board runs the TabArena splits and scores every dataset with ROC AUC, whereas TabArena scores multiclass tasks with log loss.

\textbf{Seldon-Mini and Seldon lie on the Pareto front of Elo against inference time, overall and on classification.}
Figure~\ref{fig:pareto} shows the fronts for classification and regression.
On regression, Seldon-Mini lies on the front, and Inpriora is faster than Seldon with a higher Elo.
Seldon-Mini also lies on the front of Elo against fit time, overall, on classification and on regression.

\textbf{Subsets.} TabArena labels 15 datasets as medium-sized, and Seldon ranks third on them.
On the 10 medium-sized classification datasets, Seldon ranks second among the 103 leaderboard entries rated on this subset, behind LimiX-2-1016, with an Elo of 2433.
On the 8 finance datasets, Seldon ranks in the top three on 6 and first on 2: Bank Customer Churn and credit-g.

\subsubsection{BeyondArena IID}

BeyondArena has 142 datasets with IID, temporal and grouped splits.
We report its IID track, 103 datasets with random splits, since the Seldon models do not model temporal or grouped structure.

\begin{figure}[t]
  \centering
  \includegraphics[width=\textwidth]{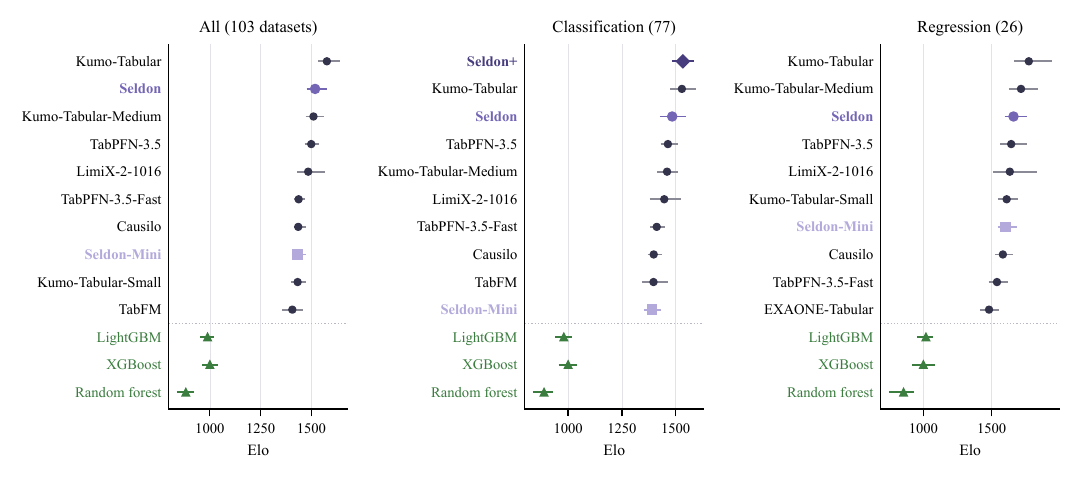}
  \caption{BeyondArena leaderboard restricted to the IID track (random splits, 103 datasets), top 10 of 42 entries of each task view. Elo with 95\% confidence interval; the default XGBoost is at 1000. Below the dotted line: the default LightGBM, XGBoost and random forest. Seldon+ is not on the official leaderboard: it is added to the classification panel with its rating from our evaluation service.}
  \label{fig:beyond}
\end{figure}

\textbf{Seldon ranks second among 42 entries on the IID track.}
Figure~\ref{fig:beyond} gives the leaderboard.
Its 42 entries cover 26 models.
Seldon reaches an Elo of 1518, 58 points below Kumo-Tabular.
It ranks second on classification and third on regression.
Seldon-Mini ranks eighth overall.
On all 142 datasets, including the temporal and grouped tracks, Seldon ranks fourth.
On the 77 IID classification datasets, our evaluation service rates Seldon+ at 1534$^{+50}_{-52}$, first among 43 entries, ahead of Kumo-Tabular (1529) and Seldon (1484).
This board runs the BeyondArena splits and metrics, and we shift its Elo so that the default XGBoost is at 1000, as on the official leaderboard; Seldon then reads 1481 on it, against 1484 officially.

\textbf{Subsets.} BeyondArena labels tables with fewer than 1{,}000 training rows as tiny, and splits features into low- and high-dimensional.
Seldon ranks second on the 17 multiclass datasets, and first on the 10 tiny, low-dimensional multiclass datasets.
Seldon-Mini ranks third on the 9 tiny, high-dimensional datasets, all tasks included.

\textbf{Seldon-Mini and Seldon lie on the Pareto fronts of Elo against fit time and against inference time.}
This holds overall, on classification and on regression.

\Needspace{16\baselineskip}
\subsubsection{TabBench-Bio}

\begin{wrapfigure}[13]{r}{0.4\textwidth}
  \vspace{-0.9cm}
  \centering
  \includegraphics[width=0.4\textwidth]{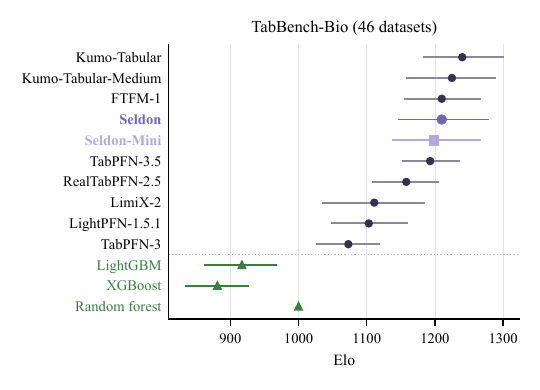}
  \vspace{-0.6cm}
  \caption{TabBench-Bio published leaderboard, top 10 of 36 entries, and the default tree-based models below the dotted line.}
  \label{fig:bio}
  \vspace{-0.5cm}
\end{wrapfigure}
TabBench-Bio \citep{kreuer2026tabbenchbio} is a living benchmark of high-dimensional biomedical tables, developed by an independent group and unrelated to our TabBench despite the similar name.
It ranks models at its reference setting, with at most 10{,}000 features and 100 training rows per dataset.
Its Elo is fitted on per-split comparisons, with the default random forest at 1000.
We use its published leaderboard of 13 November 2026, which has 36 entries and does not include Seldon+; Figure~\ref{fig:bio} gives the top 10.
Seldon+ ranks first in our own evaluation at this setting, outside the published leaderboard.
On the published leaderboard, Kumo-Tabular ranks first with an Elo of 1240, and Kumo-Tabular-Medium second with 1225.
Seldon ranks third with 1210, tied with FTFM-1, which is evaluated on the classification datasets only.
Seldon-Mini ranks fifth with 1199, and TabPFN-3.5 sixth with 1193.
The 95\% intervals of these six entries overlap.
Kumo-Tabular, Seldon and Seldon-Mini select at most 500 features by default, and the benchmark removes this limit.

\subsection{Neuralk Private Benchmarks}
\label{sec:private}

TabBench and Breaking TFMs are built and run by Neuralk.
The datasets of TabBench are public OpenML datasets, but the selection, the protocol and the leaderboard are ours and are not public; Appendices~\ref{app:tabbench} and~\ref{app:breaking-tfms} give what is needed to rebuild both benchmarks.
Every model is run by our evaluation service with the same protocol, Seldon+ included on TabBench.

\subsubsection{TabBench}

\textbf{Datasets.} TabBench holds 124 classification datasets from OpenML.
Appendix~\ref{app:tabbench} lists the datasets of the three suites and gives the full protocol.
The datasets have 500 to 150{,}000 rows, with a median of 3{,}996, and 2 to 1{,}776 features, with a median of 16.5.
Of the 124 tasks, 86 are binary and 38 multiclass, with up to 100 classes.
TabBench-Indus holds 54 industrial use cases built from public datasets, grouped by business vertical.
A dataset enters TabBench-Indus when its prediction can inform a business decision and gradient-boosted trees do not already score close to the maximum.
TabBench-Missing holds the 21 TabBench datasets with at least one missing cell, where the share of missing cells ranges from 0.01\% to 63\%.

\begin{figure}[t]
  \centering
  \includegraphics[width=\textwidth]{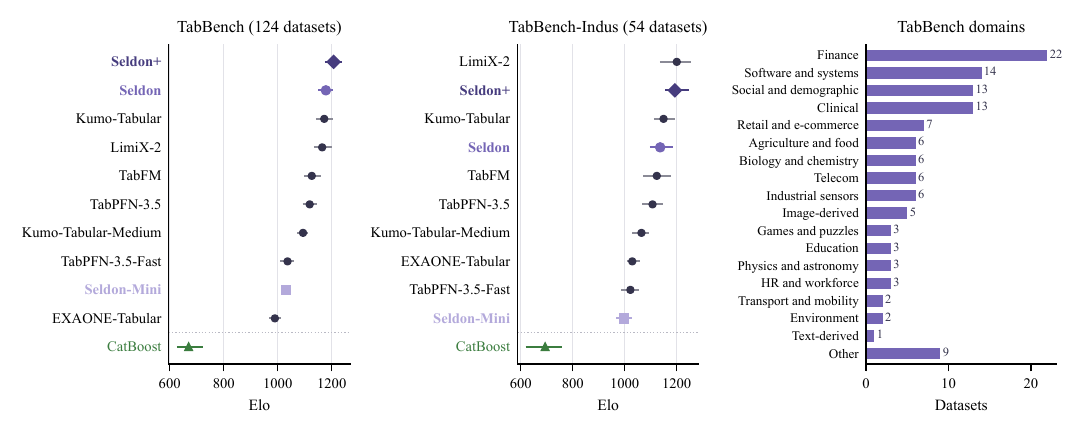}
  \caption{Left and centre: TabBench and TabBench-Indus, top 10 of 15 models per suite. Elo with 95\% confidence interval. Below the dotted line: CatBoost, the only single tree-based model on these boards. A failure takes the score of CatBoost. Right: number of TabBench datasets per domain.}
  \label{fig:tabbench}
\end{figure}

\textbf{Seldon+ and Seldon rank first and second on TabBench.}
Figure~\ref{fig:tabbench} gives TabBench and TabBench-Indus, and the domains of the TabBench datasets.
On TabBench, Seldon+ reaches an Elo of 1207 and Seldon 1178.
The 95\% intervals of the top four models overlap.
On TabBench-Indus, LimiX-2 ranks first, Seldon+ second and Seldon fourth.
On TabBench-Missing, Seldon+ and Seldon rank first and second.
Seldon-Mini ranks ninth or tenth on the three suites.

\textbf{Head-to-head.} Seldon+ wins against Seldon on 66\% of TabBench datasets and 72\% of TabBench-Indus datasets.
Against Seldon-Mini, Seldon+ wins on 81\% to 93\% of the datasets of each suite, and Seldon wins on 81\% to 86\%.

\textbf{Seldon+ ranks first in 11 of the 18 dataset groups shown.}
We place each Seldon model per group, for groups with at least five datasets.
On TabBench, Seldon+ ranks first on finance (22 datasets), social and demographic, retail and e-commerce, agriculture and food, telecom, and image-derived data.
On TabBench-Indus, it ranks first on credit, diagnosis, commerce and marketing.
On TabBench-Missing, it ranks first on datasets with under 1\% of cells missing.
Seldon ranks first on biology and chemistry, on the group labelled other, and on datasets with 1\% to 10\% of cells missing.
LimiX-2 ranks first on clinical data, industrial sensors and the software use case of TabBench-Indus.
Kumo-Tabular ranks first on software and systems.

\textbf{Seldon+ ranks first on most size, feature and class subsets.}
We also split TabBench by number of rows, features and classes.
Seldon+ ranks first on 6 of the 11 subsets, and Seldon on 3.
Seldon ranks first on the 29 datasets with fewer than 10 features.
It also ranks first on the 9 datasets with at least 50{,}000 rows and on the 5 datasets with more than 10 classes.
On these two subsets, Seldon+ runs out of memory on two and three datasets, and Kumo-Tabular and five other models fail on the same two large datasets.

\subsubsection{Breaking TFMs}

\begin{figure}[!htb]
  \centering
  \includegraphics[width=\textwidth]{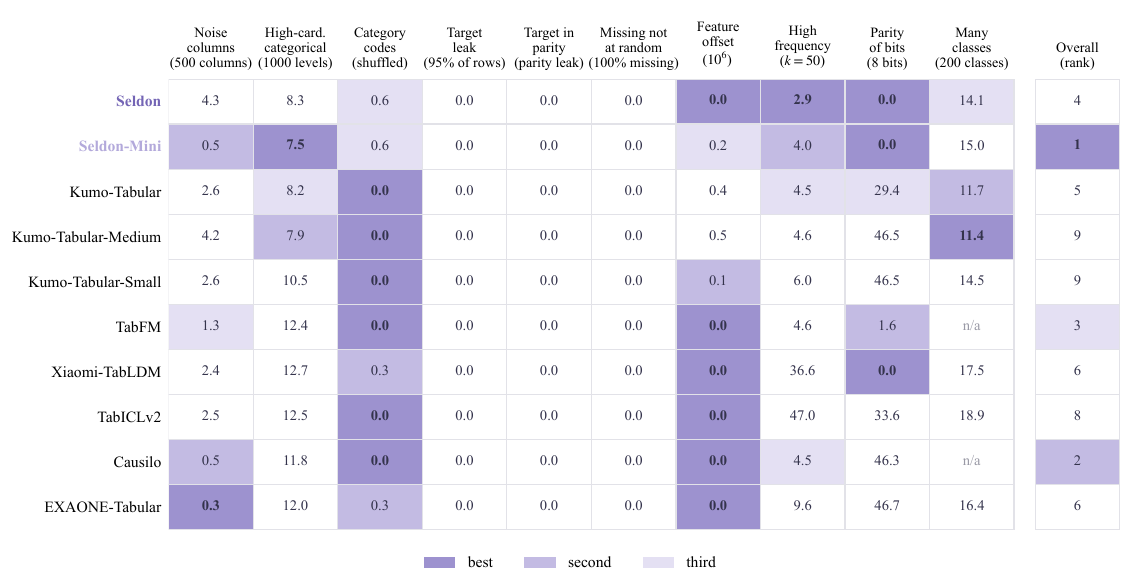}
  \caption{Breaking TFMs: relative loss (\%) of each model on each probe, defined in the text; the hardest level is in parentheses. Shading marks the three best models per column, darkest first. n/a: no score. Overall: rank by mean rank over the nine probes every model is scored on.}
  \label{fig:probes}
\end{figure}

Breaking TFMs has 10 probes, each testing one failure mode at 2 to 4 levels.
Each level is one synthetic classification table of 2{,}000 rows, and the lowest level of a probe is a control without the failure mode.
The tables are drawn from known class probabilities, so the accuracy of the Bayes-optimal classifier is known.
Appendix~\ref{app:breaking-tfms} defines every probe and level.
Our evaluation service scores each model by accuracy, with 5-fold cross-validation.
We divide the accuracy of a model on each table by the Bayes-optimal accuracy of that table.
The relative loss of a model on a probe is the drop of this ratio from the control to the hardest level, in percent.
Figure~\ref{fig:probes} gives the relative losses, with a negative loss shown as 0.
The overall rank of a model is its mean rank over the nine probes every model is scored on.
Seldon-Mini ranks first overall and Seldon fourth.
Seldon+ and the tree-based models are not evaluated on this suite.

\textbf{Seldon, Seldon-Mini and Xiaomi-TabLDM solve the parity of 8 bits.}
The target is the parity of 8 binary features, with 5\% of the labels flipped.
Seldon, Seldon-Mini and Xiaomi-TabLDM reach the Bayes-optimal accuracy of 95.4\%, and TabFM reaches 93.8\%.
Kumo-Tabular reaches 67.4\%, and the other models stay between 50.8\% and 63.3\%.

\textbf{Seldon has the highest accuracy on the high-frequency target.}
The target is the sign of $\sin(50x)$, with 5\% of the labels flipped.
Seldon reaches 91.7\% and Seldon-Mini 90.6\%, against 94.5\% for the Bayes-optimal classifier.
TabICLv2 and Xiaomi-TabLDM fall to 50.0\% and 59.8\%.

\textbf{Every model loses accuracy on many classes and on a high-cardinality categorical.}
With 200 classes, the relative loss of every model is between 11.4\% and 18.9\%.
Kumo-Tabular-Medium and Kumo-Tabular are the most accurate, with 67.5\% and 67.3\%, and Seldon reaches 65.4\%.
With a categorical feature of 1000 levels, the four best models are Seldon-Mini, Kumo-Tabular-Medium, Kumo-Tabular and Seldon, between 74.2\% and 74.5\%, against 81.0\% for the Bayes-optimal classifier.
With 500 noise columns, the relative loss is 4.3\% for Seldon and 0.5\% for Seldon-Mini.
On the target leak, the target hidden in a parity, the missing-not-at-random column, the shuffled category codes and the offset of $10^6$, the relative loss of every model is below 1\%.

\subsection{Pairwise Win Rates}

\begin{figure}[!htb]
  \centering
  \includegraphics[width=\textwidth]{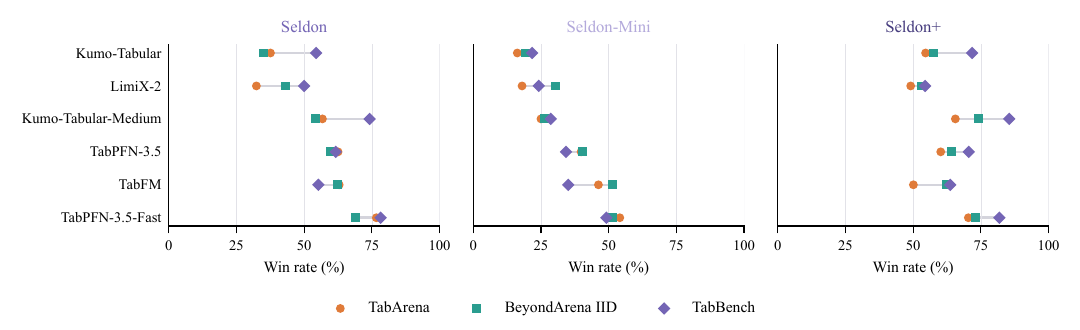}
  \caption{Win rates of each Seldon model against six leading foundation models on TabArena (51 datasets, per-split comparisons), the BeyondArena IID track (103 datasets, per-split comparisons) and TabBench (124 datasets, per-dataset comparisons). The dashed line is 50\%. LimiX-2 is the LimiX-2-1016 checkpoint on TabArena and BeyondArena, and the first LimiX-2 checkpoint on TabBench. Seldon+ is not on the TabArena and BeyondArena leaderboards; its win rates there come from our own runs of TabArena classification (38 datasets, ROC AUC) and BeyondArena IID classification (77 datasets), with per-split comparisons.}
  \label{fig:winrates}
\end{figure}

\textbf{Seldon wins more than half of its comparisons against at least four of the six leading models on each benchmark.}
Figure~\ref{fig:winrates} gives the win rates of each Seldon model against six leading foundation models.
Seldon wins 54\% to 77\% of comparisons against Kumo-Tabular-Medium, TabPFN-3.5, TabFM and TabPFN-3.5-Fast.
It wins 35\% to 38\% against Kumo-Tabular on TabArena and BeyondArena, and 54\% on TabBench.
Against LimiX-2, Seldon wins 32\% on TabArena, 43\% on BeyondArena and 50\% on TabBench.
Seldon-Mini wins 54\% and 51\% against TabPFN-3.5-Fast on TabArena and BeyondArena, and 49\% on TabBench.
Seldon+ wins 54\% to 86\% against all six models on TabBench, and 53\% to 74\% on BeyondArena IID classification.
On TabArena classification, it wins 50\% to 70\% against five of them, and 49\% against LimiX-2.

\FloatBarrier

\section{Conclusion}
\label{sec:conclusion}

This report introduced Seldon, a family of three tabular foundation models.
Seldon-Mini and Seldon are trained on synthetic tables only and predict on a new table in a single forward pass, with no training and no tuning.
On the public benchmarks, Seldon ranks fourth among 109 entries on TabArena and second among 42 on the IID track of BeyondArena, and Seldon-Mini and Seldon both lie on the TabArena Pareto front of Elo against inference time.
On our own benchmarks, Seldon+ and Seldon rank first and second on TabBench, and Breaking TFMs shows that Seldon and Seldon-Mini solve the parity of eight bits, which most of the other foundation models we evaluate do not.
A small model can remain competitive: with 27M parameters, Seldon-Mini lies on that front alongside Seldon, eight times larger.

\enlargethispage{2\baselineskip}
\paragraph{The Seldon Plan.}
A single table of independent rows is only the beginning.
The structured data on which decisions are made is richer: rows arrive over time, tables are linked to one another, cells hold text and images next to numbers, and the questions asked of the data are often causal, about the effect of an action rather than the value of a label.
Synthetic priors are a natural path towards a broader array of capabilities, since structural causal models already describe time, relations and interventions.
Seldon is the first step on a journey.

\clearpage
\bibliographystyle{plainnat}
\bibliography{references}

\section*{Contributors}
\addcontentsline{toc}{section}{Contributors}

Names are listed in alphabetical order of last name.

\textbf{Research.}
Stanislas Angebault,
Nicolas Atienza,
Nicola Cancedda,
Aziz Fouché,
Salomé Gobbi,
Ekaterina Iakovleva,
Nicolas Makaroff,
Arthur Nahmias,
Paul Novello,
Alexandre Pasquiou.

\textbf{Operations, go-to-market, marketing and engineering.}
Jeremy Ben Sadoun,
Frédéric Catherine,
Raphael Ducasse,
Marie Llorca,
Theo Marcolini,
Paul Martineau,
Antoine Moissenot,
Quentin Navellou,
Aaron Stillwell,
Alexis Wychovaniak.

\appendix

\section{TabBench}
\label{app:tabbench}

TabBench is not public. This appendix gives what is needed to rebuild it: the datasets of each suite, how a table is read, the splits, the metric and the scoring rules. Tables~\ref{tab:tabbench-datasets} to~\ref{tab:tabbench-missing-datasets} list the datasets.

\subsection{Suites}

TabBench has three suites. All three are classification suites, and every dataset is an OpenML dataset, named by its OpenML ID.

\begin{itemize}
  \item \textbf{TabBench} holds 124 datasets, chosen to cover as many kinds of real tabular data as possible (Table~\ref{tab:tabbench-datasets}). Each dataset has one domain label, used only to group results. The datasets have 500 to 150{,}000 rows, with a median of 3{,}996, and 2 to 1{,}776 features, with a median of 16.5. Of the 124 tasks, 86 are binary and 38 multiclass, with up to 100 classes.
  \item \textbf{TabBench-Indus} holds 54 datasets, each an industrial use case built from a public dataset (Table~\ref{tab:tabbench-indus-datasets}). A dataset enters the suite when it is relevant to a business, when its prediction can inform a decision, and when a tabular model is the right tool for it. Duplicates are excluded, and so are datasets on which gradient-boosted trees already score close to the maximum. Each dataset has one use-case label. Of the 54 datasets, 52 are also in TabBench; the other two are KDDCup09\_churn (1112) and Credit-Risk-Dataset (43454).
  \item \textbf{TabBench-Missing} holds the 21 TabBench datasets with at least one missing cell (Table~\ref{tab:tabbench-missing-datasets}). The share of missing cells ranges from 0.01\% to 63\%. The datasets are grouped in three bands: under 1\% of cells missing (10 datasets), 1\% to 10\% (8) and over 10\% (3). TabBench-Missing needs no run of its own: its scores are the TabBench scores of these 21 datasets.
\end{itemize}

\subsection{Data}

Each table is the full OpenML dataset, as returned by \texttt{openml.datasets.get\_dataset(id).get\_data()}. No row is removed or subsampled, and no feature is removed. The target is the default target attribute of the OpenML dataset, and \texttt{loan\_status} for Credit-Risk-Dataset (43454), which has none; every other column is a feature. Our evaluation service reads a parquet snapshot of each table, taken once, so that every run sees the same rows.

The service hands each model the raw features of the training rows. It replaces $\pm\infty$ with a missing value and drops the columns that are missing on every training row. Missing values are kept. A model that does not encode categorical or datetime features itself receives a default encoding, fitted on the training rows only: constant and duplicate columns are dropped, a datetime becomes seconds since the epoch, plus its year, month, day and weekday when these vary over the training rows, and every other non-numeric column becomes integer codes in sorted order of its levels (the declared order for a category column), with a level unseen in training coded as missing. The Seldon models receive the raw features and apply their own preprocessing.

\subsection{Protocol}

\paragraph{Splits.} Each dataset is evaluated with 5-fold cross-validation, stratified on the target and shuffled with seed~0. The folds are those of scikit-learn's \texttt{StratifiedKFold}, with \texttt{n\_splits=5}, \texttt{shuffle=True} and \texttt{random\_state=0}, applied to the rows in their OpenML order. A model is fitted on four folds and predicts class probabilities on the fifth, once per fold.

\paragraph{Metric.} Each fold is scored with ROC AUC. For a binary task, it is the ROC AUC of the second class in sorted label order. For a multiclass task, it is the mean of the one-vs-rest ROC AUCs over the classes present in the test fold. The score of a model on a dataset is the mean over the five folds.

\paragraph{Failures.} A model fails on a dataset when any of its five folds raises an error, runs out of memory, or exceeds the time budget of 9{,}000 seconds per dataset, that is 1{,}800 seconds per fold. A failed dataset receives the score of the default CatBoost on that dataset.

\paragraph{Ranking.} Models are ranked with the Elo and the win rates of Section~\ref{sec:protocol}. The external models are first rated among themselves, with 100 bootstrap resamples over datasets for the confidence intervals. Each Seldon model is then rated against this fixed field.

\subsection{Datasets}

\begingroup\scriptsize\setlength{\tabcolsep}{3pt}
\begin{longtable}{@{}r p{6.2cm} r r r p{3.6cm}@{}}
\caption{TabBench: the 124 OpenML datasets, with their number of rows, features (target excluded) and classes, and the domain used to group them.}\label{tab:tabbench-datasets}\\
\toprule
\textbf{ID} & \textbf{Name} & \textbf{Rows} & \textbf{Feat.} & \textbf{Cl.} & \textbf{Domain} \\
\midrule
\endfirsthead
\multicolumn{6}{@{}l}{\tablename~\thetable{} (continued)}\\
\toprule
\textbf{ID} & \textbf{Name} & \textbf{Rows} & \textbf{Feat.} & \textbf{Cl.} & \textbf{Domain} \\
\midrule
\endhead
\bottomrule
\endfoot
18 & mfeat-morphological & 2{,}000 & 6 & 10 & image-derived \\
22 & mfeat-zernike & 2{,}000 & 47 & 10 & image-derived \\
23 & cmc & 1{,}473 & 9 & 3 & social and demographic \\
29 & credit-approval & 690 & 15 & 2 & finance \\
31 & credit-g & 1{,}000 & 20 & 2 & finance \\
37 & diabetes & 768 & 8 & 2 & clinical \\
54 & vehicle & 846 & 18 & 4 & image-derived \\
60 & waveform-5000 & 5{,}000 & 40 & 3 & other \\
137 & BNG(tic-tac-toe) & 39{,}366 & 9 & 2 & games and puzzles \\
179 & adult & 48{,}842 & 14 & 2 & social and demographic \\
181 & yeast & 1{,}484 & 8 & 10 & biology and chemistry \\
185 & baseball & 1{,}340 & 16 & 3 & other \\
188 & eucalyptus & 736 & 19 & 5 & agriculture and food \\
255 & BNG(cmc) & 55{,}296 & 9 & 3 & social and demographic \\
301 & ozone\_level & 2{,}536 & 72 & 2 & environment \\
310 & mammography & 11{,}183 & 6 & 2 & clinical \\
335 & monks-problems-3 & 554 & 6 & 2 & other \\
372 & internet\_usage & 10{,}108 & 71 & 46 & telecom \\
469 & analcatdata\_dmft & 797 & 4 & 6 & clinical \\
741 & rmftsa\_sleepdata & 1{,}024 & 2 & 2 & clinical \\
803 & delta\_ailerons & 7{,}129 & 5 & 2 & transport and mobility \\
930 & colleges\_usnews & 1{,}302 & 33 & 2 & education \\
1037 & ada\_prior & 4{,}562 & 14 & 2 & social and demographic \\
1043 & ada\_agnostic & 4{,}562 & 48 & 2 & social and demographic \\
1046 & mozilla4 & 15{,}545 & 5 & 2 & software and systems \\
1049 & pc4 & 1{,}458 & 37 & 2 & software and systems \\
1050 & pc3 & 1{,}563 & 37 & 2 & software and systems \\
1063 & kc2 & 522 & 21 & 2 & software and systems \\
1067 & kc1 & 2{,}109 & 21 & 2 & software and systems \\
1068 & pc1 & 1{,}109 & 21 & 2 & software and systems \\
1120 & MagicTelescope & 19{,}020 & 10 & 2 & physics and astronomy \\
1220 & Click\_prediction\_small & 39{,}948 & 9 & 2 & retail and e-commerce \\
1444 & PizzaCutter3 & 1{,}043 & 37 & 2 & software and systems \\
1464 & blood-transfusion-service-center & 748 & 4 & 2 & clinical \\
1475 & first-order-theorem-proving & 6{,}118 & 51 & 6 & software and systems \\
1480 & ilpd & 583 & 10 & 2 & clinical \\
1487 & ozone-level-8hr & 2{,}534 & 72 & 2 & environment \\
1489 & phoneme & 5{,}404 & 5 & 2 & other \\
1494 & qsar-biodeg & 1{,}055 & 41 & 2 & biology and chemistry \\
1509 & walking-activity & 149{,}332 & 4 & 22 & industrial sensors \\
1548 & autoUniv-au4-2500 & 2{,}500 & 100 & 3 & other \\
1552 & autoUniv-au7-1100 & 1{,}100 & 12 & 5 & other \\
1557 & abalone & 4{,}177 & 8 & 3 & agriculture and food \\
4538 & GesturePhaseSegmentationProcessed & 9{,}873 & 32 & 5 & industrial sensors \\
6332 & cylinder-bands & 540 & 37 & 2 & industrial sensors \\
23381 & dresses-sales & 500 & 12 & 2 & retail and e-commerce \\
40478 & thyroid-dis & 2{,}800 & 26 & 5 & clinical \\
40646 & GAMETES\_Epistasis\_2-Way\_20atts\_0.1H\_EDM-1\_1 & 1{,}600 & 20 & 2 & biology and chemistry \\
40649 & GAMETES\_Heterogeneity\_20atts\_1600\_\allowbreak Het\_0.4\_0.2\_50\_EDM-2\_001 & 1{,}600 & 20 & 2 & biology and chemistry \\
40668 & connect-4 & 67{,}557 & 42 & 3 & games and puzzles \\
40677 & led24 & 3{,}200 & 24 & 10 & image-derived \\
40678 & led7 & 3{,}200 & 7 & 10 & image-derived \\
40701 & churn & 5{,}000 & 20 & 2 & telecom \\
40981 & Australian & 690 & 14 & 2 & finance \\
40982 & steel-plates-fault & 1{,}941 & 27 & 7 & industrial sensors \\
40994 & climate-model-simulation-crashes & 540 & 18 & 2 & software and systems \\
41027 & jungle\_chess\_2pcs\_raw\_endgame\_complete & 44{,}819 & 6 & 3 & games and puzzles \\
41143 & jasmine & 2{,}984 & 144 & 2 & text-derived \\
41156 & ada & 4{,}147 & 48 & 2 & social and demographic \\
41169 & helena & 65{,}196 & 27 & 100 & other \\
41671 & microaggregation2 & 20{,}000 & 20 & 5 & social and demographic \\
41705 & ASP-POTASSCO-classification & 1{,}294 & 142 & 11 & software and systems \\
41721 & FOREX\_audcad-day-High & 1{,}834 & 11 & 2 & finance \\
41865 & FOREX\_audsgd-hour-High & 43{,}825 & 11 & 2 & finance \\
41875 & FOREX\_audchf-day-High & 1{,}833 & 11 & 2 & finance \\
41882 & FOREX\_audjpy-day-High & 1{,}832 & 11 & 2 & finance \\
42178 & telco-customer-churn & 7{,}043 & 19 & 2 & telecom \\
42464 & Waterstress & 1{,}188 & 22 & 2 & agriculture and food \\
43582 & Pima-Indians-Diabetes & 768 & 8 & 2 & clinical \\
43672 & Heart-Disease-Dataset-(Comprehensive) & 1{,}190 & 11 & 2 & clinical \\
43812 & Performance-Prediction & 1{,}340 & 20 & 2 & other \\
44089 & credit & 16{,}714 & 10 & 2 & finance \\
44123 & house\_16H & 13{,}488 & 16 & 2 & social and demographic \\
44124 & kdd\_ipums\_la\_97-small & 5{,}188 & 20 & 2 & social and demographic \\
44130 & eye\_movements & 7{,}608 & 20 & 2 & industrial sensors \\
44150 & VulNoneVul & 5{,}692 & 16 & 2 & software and systems \\
44160 & rl & 4{,}970 & 12 & 2 & other \\
44162 & compass & 16{,}644 & 17 & 2 & social and demographic \\
44186 & KDDCup09\_upselling & 5{,}128 & 49 & 2 & telecom \\
44528 & airlines\_seed\_0\_nrows\_2000\_\allowbreak nclasses\_10\_ncols\_100\_stratify\_True & 2{,}000 & 7 & 2 & transport and mobility \\
44966 & solar\_flare & 1{,}066 & 10 & 8 & physics and astronomy \\
44967 & student\_performance\_por & 649 & 30 & 17 & education \\
44971 & white\_wine & 4{,}898 & 11 & 7 & agriculture and food \\
44972 & red\_wine & 1{,}599 & 11 & 6 & agriculture and food \\
45052 & cmc & 1{,}473 & 9 & 3 & social and demographic \\
45062 & shrutime & 10{,}000 & 10 & 2 & finance \\
45064 & Insurance & 23{,}548 & 10 & 2 & finance \\
45067 & okcupid\_stem & 26{,}677 & 13 & 3 & social and demographic \\
45075 & KDD & 5{,}032 & 45 & 2 & telecom \\
45547 & Cardiovascular-Disease-dataset & 70{,}000 & 11 & 2 & clinical \\
45553 & FICO-HELOC-cleaned & 9{,}871 & 23 & 2 & finance \\
45558 & Pulsar-Dataset-HTRU2 & 17{,}898 & 8 & 2 & physics and astronomy \\
45578 & California-Housing-Classification & 20{,}640 & 8 & 2 & social and demographic \\
46356 & GermanCreditData & 1{,}000 & 21 & 2 & finance \\
46441 & Credit\_Score\_Classification & 100{,}000 & 27 & 3 & finance \\
46905 & Amazon\_employee\_access & 32{,}769 & 9 & 2 & HR and workforce \\
46910 & bank-marketing & 45{,}211 & 13 & 2 & finance \\
46911 & Bank\_Customer\_Churn & 10{,}000 & 10 & 2 & finance \\
46912 & Bioresponse & 3{,}751 & 1{,}776 & 2 & biology and chemistry \\
46916 & coil2000\_insurance\_policies & 9{,}822 & 85 & 2 & finance \\
46919 & credit\_card\_clients\_default & 30{,}000 & 23 & 2 & finance \\
46922 & Diabetes130US & 71{,}518 & 47 & 2 & clinical \\
46924 & E-CommereShippingData & 10{,}999 & 10 & 2 & retail and e-commerce \\
46925 & Employee & 4{,}653 & 8 & 2 & HR and workforce \\
46927 & Fitness\_Club & 1{,}500 & 6 & 2 & retail and e-commerce \\
46929 & GiveMeSomeCredit & 150{,}000 & 10 & 2 & finance \\
46932 & heloc & 10{,}459 & 23 & 2 & finance \\
46933 & hiva\_agnostic & 3{,}845 & 1{,}617 & 3 & biology and chemistry \\
46935 & HR\_Analytics\_Job\_Change\_of\_Data\_Scientists & 19{,}158 & 12 & 2 & HR and workforce \\
46937 & in\_vehicle\_coupon\_recommendation & 12{,}684 & 24 & 2 & retail and e-commerce \\
46938 & Is-this-a-good-customer & 1{,}723 & 13 & 2 & finance \\
46939 & kddcup09\_appetency & 50{,}000 & 212 & 2 & telecom \\
46940 & Marketing\_Campaign & 2{,}240 & 25 & 2 & retail and e-commerce \\
46941 & maternal\_health\_risk & 1{,}014 & 6 & 3 & clinical \\
46947 & online\_shoppers\_intention & 12{,}330 & 17 & 2 & retail and e-commerce \\
46950 & polish\_companies\_bankruptcy & 5{,}910 & 64 & 2 & finance \\
46951 & Pumpkin\_Seeds & 2{,}500 & 12 & 2 & agriculture and food \\
46956 & seismic-bumps & 2{,}584 & 15 & 2 & industrial sensors \\
46960 & students\_dropout\_and\_academic\_success & 4{,}424 & 36 & 3 & education \\
46962 & taiwanese\_bankruptcy\_prediction & 6{,}819 & 94 & 2 & finance \\
46963 & website\_phishing & 1{,}353 & 9 & 3 & software and systems \\
46969 & NATICUSdroid & 7{,}491 & 86 & 2 & software and systems \\
46979 & jm1 & 10{,}885 & 21 & 2 & software and systems \\
46980 & MIC & 1{,}699 & 111 & 8 & clinical \\
\end{longtable}
\endgroup

\begingroup\scriptsize\setlength{\tabcolsep}{3pt}
\begin{longtable}{@{}r p{6.2cm} r r r p{3.6cm}@{}}
\caption{TabBench-Indus: the 54 OpenML datasets and the business use case used to group them. All but KDDCup09\_churn (1112) and Credit-Risk-Dataset (43454) are also in TabBench.}\label{tab:tabbench-indus-datasets}\\
\toprule
\textbf{ID} & \textbf{Name} & \textbf{Rows} & \textbf{Feat.} & \textbf{Cl.} & \textbf{Use case} \\
\midrule
\endfirsthead
\multicolumn{6}{@{}l}{\tablename~\thetable{} (continued)}\\
\toprule
\textbf{ID} & \textbf{Name} & \textbf{Rows} & \textbf{Feat.} & \textbf{Cl.} & \textbf{Use case} \\
\midrule
\endhead
\bottomrule
\endfoot
37 & diabetes & 768 & 8 & 2 & diagnosis \\
179 & adult & 48{,}842 & 14 & 2 & marketing \\
310 & mammography & 11{,}183 & 6 & 2 & diagnosis \\
1046 & mozilla4 & 15{,}545 & 5 & 2 & software \\
1067 & kc1 & 2{,}109 & 21 & 2 & software \\
1112 & KDDCup09\_churn & 50{,}000 & 230 & 2 & churn \\
1220 & Click\_prediction\_small & 39{,}948 & 9 & 2 & marketing \\
1480 & ilpd & 583 & 10 & 2 & diagnosis \\
6332 & cylinder-bands & 540 & 37 & 2 & sensor \\
23381 & dresses-sales & 500 & 12 & 2 & commerce \\
40478 & thyroid-dis & 2{,}800 & 26 & 5 & diagnosis \\
40701 & churn & 5{,}000 & 20 & 2 & churn \\
40981 & Australian & 690 & 14 & 2 & credit \\
40982 & steel-plates-fault & 1{,}941 & 27 & 7 & sensor \\
40994 & climate-model-simulation-crashes & 540 & 18 & 2 & software \\
41721 & FOREX\_audcad-day-High & 1{,}834 & 11 & 2 & stocks \\
41865 & FOREX\_audsgd-hour-High & 43{,}825 & 11 & 2 & stocks \\
41875 & FOREX\_audchf-day-High & 1{,}833 & 11 & 2 & stocks \\
41882 & FOREX\_audjpy-day-High & 1{,}832 & 11 & 2 & stocks \\
42178 & telco-customer-churn & 7{,}043 & 19 & 2 & churn \\
43454 & Credit-Risk-Dataset & 32{,}581 & 11 & 2 & credit \\
43672 & Heart-Disease-Dataset-(Comprehensive) & 1{,}190 & 11 & 2 & diagnosis \\
44186 & KDDCup09\_upselling & 5{,}128 & 49 & 2 & commerce \\
44528 & airlines\_seed\_0\_nrows\_2000\_\allowbreak nclasses\_10\_ncols\_100\_stratify\_True & 2{,}000 & 7 & 2 & logistics \\
45064 & Insurance & 23{,}548 & 10 & 2 & insurance \\
45067 & okcupid\_stem & 26{,}677 & 13 & 3 & marketing \\
45075 & KDD & 5{,}032 & 45 & 2 & marketing \\
45547 & Cardiovascular-Disease-dataset & 70{,}000 & 11 & 2 & diagnosis \\
45553 & FICO-HELOC-cleaned & 9{,}871 & 23 & 2 & credit \\
46356 & GermanCreditData & 1{,}000 & 21 & 2 & credit \\
46441 & Credit\_Score\_Classification & 100{,}000 & 27 & 3 & credit \\
46905 & Amazon\_employee\_access & 32{,}769 & 9 & 2 & HR \\
46910 & bank-marketing & 45{,}211 & 13 & 2 & marketing \\
46911 & Bank\_Customer\_Churn & 10{,}000 & 10 & 2 & churn \\
46916 & coil2000\_insurance\_policies & 9{,}822 & 85 & 2 & insurance \\
46919 & credit\_card\_clients\_default & 30{,}000 & 23 & 2 & credit \\
46922 & Diabetes130US & 71{,}518 & 47 & 2 & diagnosis \\
46924 & E-CommereShippingData & 10{,}999 & 10 & 2 & logistics \\
46925 & Employee & 4{,}653 & 8 & 2 & HR \\
46927 & Fitness\_Club & 1{,}500 & 6 & 2 & commerce \\
46929 & GiveMeSomeCredit & 150{,}000 & 10 & 2 & credit \\
46935 & HR\_Analytics\_Job\_Change\_of\_Data\_Scientists & 19{,}158 & 12 & 2 & HR \\
46937 & in\_vehicle\_coupon\_recommendation & 12{,}684 & 24 & 2 & commerce \\
46938 & Is-this-a-good-customer & 1{,}723 & 13 & 2 & credit \\
46939 & kddcup09\_appetency & 50{,}000 & 212 & 2 & marketing \\
46940 & Marketing\_Campaign & 2{,}240 & 25 & 2 & commerce \\
46941 & maternal\_health\_risk & 1{,}014 & 6 & 3 & diagnosis \\
46947 & online\_shoppers\_intention & 12{,}330 & 17 & 2 & commerce \\
46950 & polish\_companies\_bankruptcy & 5{,}910 & 64 & 2 & credit \\
46956 & seismic-bumps & 2{,}584 & 15 & 2 & sensor \\
46962 & taiwanese\_bankruptcy\_prediction & 6{,}819 & 94 & 2 & credit \\
46963 & website\_phishing & 1{,}353 & 9 & 3 & software \\
46969 & NATICUSdroid & 7{,}491 & 86 & 2 & software \\
46980 & MIC & 1{,}699 & 111 & 8 & diagnosis \\
\end{longtable}
\endgroup

\begingroup\scriptsize\setlength{\tabcolsep}{3pt}
\begin{longtable}{@{}r p{6.2cm} r r r p{3.6cm}@{}}
\caption{TabBench-Missing: the 21 TabBench datasets with at least one missing cell, and the band of their share of missing cells.}\label{tab:tabbench-missing-datasets}\\
\toprule
\textbf{ID} & \textbf{Name} & \textbf{Rows} & \textbf{Feat.} & \textbf{Cl.} & \textbf{Missing cells} \\
\midrule
\endfirsthead
\multicolumn{6}{@{}l}{\tablename~\thetable{} (continued)}\\
\toprule
\textbf{ID} & \textbf{Name} & \textbf{Rows} & \textbf{Feat.} & \textbf{Cl.} & \textbf{Missing cells} \\
\midrule
\endhead
\bottomrule
\endfoot
29 & credit-approval & 690 & 15 & 2 & under 1\% of cells missing \\
179 & adult & 48{,}842 & 14 & 2 & under 1\% of cells missing \\
185 & baseball & 1{,}340 & 16 & 3 & under 1\% of cells missing \\
188 & eucalyptus & 736 & 19 & 5 & 1 to 10\% of cells missing \\
372 & internet\_usage & 10{,}108 & 71 & 46 & under 1\% of cells missing \\
930 & colleges\_usnews & 1{,}302 & 33 & 2 & over 10\% of cells missing \\
1037 & ada\_prior & 4{,}562 & 14 & 2 & under 1\% of cells missing \\
6332 & cylinder-bands & 540 & 37 & 2 & 1 to 10\% of cells missing \\
23381 & dresses-sales & 500 & 12 & 2 & over 10\% of cells missing \\
41705 & ASP-POTASSCO-classification & 1{,}294 & 142 & 11 & 1 to 10\% of cells missing \\
43812 & Performance-Prediction & 1{,}340 & 20 & 2 & under 1\% of cells missing \\
46441 & Credit\_Score\_Classification & 100{,}000 & 27 & 3 & 1 to 10\% of cells missing \\
46922 & Diabetes130US & 71{,}518 & 47 & 2 & 1 to 10\% of cells missing \\
46927 & Fitness\_Club & 1{,}500 & 6 & 2 & under 1\% of cells missing \\
46929 & GiveMeSomeCredit & 150{,}000 & 10 & 2 & 1 to 10\% of cells missing \\
46935 & HR\_Analytics\_Job\_Change\_of\_Data\_Scientists & 19{,}158 & 12 & 2 & under 1\% of cells missing \\
46939 & kddcup09\_appetency & 50{,}000 & 212 & 2 & over 10\% of cells missing \\
46940 & Marketing\_Campaign & 2{,}240 & 25 & 2 & under 1\% of cells missing \\
46950 & polish\_companies\_bankruptcy & 5{,}910 & 64 & 2 & 1 to 10\% of cells missing \\
46979 & jm1 & 10{,}885 & 21 & 2 & under 1\% of cells missing \\
46980 & MIC & 1{,}699 & 111 & 8 & 1 to 10\% of cells missing \\
\end{longtable}
\endgroup

\clearpage
\section{Breaking TFMs}
\label{app:breaking-tfms}

Breaking TFMs is a suite of 31 synthetic binary and multiclass classification tables. They form 10 probes, each testing one failure mode at 2 to 4 levels. The lowest level of each probe is its control, the same kind of table without the failure mode. Table~\ref{tab:probes} lists the probes.

\subsection{Tables}

Each table has 2{,}000 rows, and the class probabilities $p(y \mid x)$ of each row are known in closed form. A label is drawn from them, except in two cases. In the target leak, target in parity and missing not at random probes, the label is drawn from the base probability, the leak or the missingness is drawn given the label, and $p(y \mid x)$ is the posterior. In the many-classes probe, the classes are balanced and the features are drawn given the class. The tables are fixed: each is drawn once, with a random generator seeded with 0.

Most probes share the same informative features: five standard normal features $x_1, \dots, x_5$ and a logit
\begin{equation}
  \ell(x) = \sum_{j=1}^{5} w_j x_j, \qquad w_j \in \{-0.8, +0.8\} \text{ drawn uniformly},
  \label{eq:probe-logit}
\end{equation}
with $p(y = 1 \mid x) = \sigma(\ell(x))$ and $\sigma$ the logistic function. Below, ``base'' means these five features with this logit. A probe adds a column to the base, modifies the base, or replaces the base.

\begin{description}
  \item[Noise columns.] The base plus $k \in \{0, 10, 100, 500\}$ standard normal columns, independent of the target.
  \item[High-cardinality categorical.] The base plus one categorical column of $L \in \{10, 100, 1000\}$ equally likely levels, stored as strings. Level $c$ adds $b_c \sim \mathcal{N}(0, 1.5^2)$ to the logit: $p(y = 1 \mid x) = \sigma(\ell(x) + b_c)$.
  \item[Category codes.] The base plus one categorical column of 20 equally likely levels. Level $c \in \{0, \dots, 19\}$ adds $-2 + 4c/19$ to the logit, a linear ramp from $-2$ to $2$. The column is stored as integers. In the control, the integer is $c$, so the logit is monotone in it. At the hardest level, the integer is $\pi(c)$ for a fixed random permutation $\pi$ of the 20 codes.
  \item[Target leak.] The base plus 51 further columns: 50 standard normal noise columns and, at a random position among them, one binary leak column. With probability $q \in \{0, 0.8, 0.95\}$ the leak equals $y$, else it is a fair coin. The class probabilities are the posterior given the base and the leak, $p(y \mid x) \propto p_0(y \mid x) \, P(\text{leak} \mid y)$, where $p_0$ is the base probability, with $P(\text{leak} = y \mid y) = (1+q)/2$.
  \item[Target in parity.] As the target leak, with an integer leak column $2u + r$, where $u$ is uniform over $\{0, \dots, 19\}$. In the control, $r$ is a fair coin; at the hardest level, $r = y$, so the parity of the column is the target and the Bayes-optimal accuracy is 100\%.
  \item[Missing not at random.] The base plus one standard normal column, independent of the target, made missing on a share $m \in \{0, 0.3, 1\}$ of the rows with $y = 1$ and on no row with $y = 0$. The class probabilities are the posterior given the base and whether the column is missing: a missing value implies $y = 1$.
  \item[Feature offset.] The base, with a constant $10^s$, $s \in \{0, 6\}$, added to each of the five features. In 32-bit floats, the features lose precision as $s$ grows. The class probabilities are those of the base.
  \item[High frequency.] One feature $x_0$, uniform over $[-\pi, \pi]$, and four standard normal noise features. $p(y = 1 \mid x) = 0.95$ when $\sin(k x_0) > 0$ and $0.05$ otherwise, for $k \in \{1, 5, 20, 50\}$, so 5\% of the labels are flipped.
  \item[Parity of bits.] Eight uniform binary features $b_1, \dots, b_8$. $p(y = 1 \mid x) = 0.95$ when $b_1 + \dots + b_n$ is odd and $0.05$ otherwise, for $n \in \{1, 4, 6, 8\}$.
  \item[Many classes.] $K \in \{2, 10, 50, 200\}$ classes, each with $2000/K$ rows. Each class $k$ has a centre $\mu_k \sim \mathcal{N}(0, 1.5^2 I_{10})$, and a row of class $y$ has the ten features $x \sim \mathcal{N}(\mu_y, I_{10})$. The class probabilities are $p(y = k \mid x) \propto \exp(-\lVert x - \mu_k \rVert^2 / 2)$.
\end{description}

\subsection{Protocol}

Each table is evaluated with the splits and preprocessing of TabBench (Appendix~\ref{app:tabbench}): 5-fold stratified cross-validation with seed~0. Unlike on TabBench, a failure is not replaced by the score of CatBoost: the model has no score on that table. Each fold is scored with accuracy, the share of test rows whose most probable class is the label, and the score of a table is the mean over its five folds.

The Bayes-optimal classifier predicts $\arg\max_y p(y \mid x)$. Its accuracy $a^\star$ is computed on all 2{,}000 rows of the table, from the drawn labels, and Table~\ref{tab:probes} gives it for every level. For a model of accuracy $a$ on a table, the normalised accuracy is $a / a^\star$. The relative loss of a model on a probe is the drop of the normalised accuracy from the control to the hardest level, in percent:
\begin{equation}
  \Delta = 100 \left( \frac{a_{\text{control}}}{a^\star_{\text{control}}} - \frac{a_{\text{hardest}}}{a^\star_{\text{hardest}}} \right).
  \label{eq:relative-loss}
\end{equation}
A negative relative loss is shown as 0 in Figure~\ref{fig:probes}. The overall position of a model is its mean rank over the nine probes on which every model is scored, all but many classes, with tied models taking their average rank.

\begin{table}[t]
  \centering
  \small
  \caption{The probes of Breaking TFMs. Features: number of feature columns, from the control to the hardest level where it changes. Levels: the level of each table, the control first. $a^\star$: accuracy of the Bayes-optimal classifier on each table, in percent, in the same order.}
  \label{tab:probes}
  \begin{tabular}{@{}l l l l@{}}
    \toprule
    \textbf{Probe} & \textbf{Features} & \textbf{Levels} & $\boldsymbol{a^\star}$ \textbf{(\%)} \\
    \midrule
    Noise columns & 5 to 505 & 0, 10, 100, 500 columns & 76.7, 76.7, 76.7, 76.7 \\
    High-card.\ categorical & 6 & 10, 100, 1000 levels & 77.8, 79.8, 81.0 \\
    Category codes & 6 & ordered, shuffled & 79.1, 79.1 \\
    Target leak & 56 & $q = 0, 0.8, 0.95$ & 76.7, 89.8, 97.2 \\
    Target in parity & 56 & coin flip, target & 76.7, 100.0 \\
    Missing not at random & 6 & 0\%, 30\%, 100\% missing & 76.7, 79.8, 100.0 \\
    Feature offset & 5 & $10^0, 10^6$ & 76.7, 76.7 \\
    High frequency & 5 & $k = 1, 5, 20, 50$ & 94.2, 94.5, 94.5, 94.5 \\
    Parity of bits & 8 & 1, 4, 6, 8 bits & 95.2, 94.8, 94.1, 95.4 \\
    Many classes & 10 & 2, 10, 50, 200 classes & 100.0, 96.4, 91.8, 76.2 \\
    \bottomrule
  \end{tabular}
\end{table}

\clearpage

\end{document}